# Online search of unknown terrains using a dynamical system-based path planning approach

Karan Sridharan
Department of Mechanical Engineering
Wichita State University
Wichita, Kansas 67260
Email: kxsridharan@shockers.wichita.edu

Patrick McNamee
Department of Mechanical Engineering
San Diego State University
San Diego, California 92182
Email: pmcnamee5123@sdsu.edu

Zahra Nili Ahmadabadi[1]
Department of Mechanical Engineering
San Diego State University
San Diego, California 92182
Email: zniliahmadabadi@sdsu.edu
ORCID ID: 0000-0003-3226-4425

Jeffrey Hudack
Information Directorate
Air Force Research Laboratory
Rome, NY 13441
Email: jeffrey.hudack@us.af.mil

**ABSTRACT**

Surveillance and exploration of large environments is a tedious task. In spaces with limited environmental cues, random-like search is an effective approach as it allows the robot to perform online coverage of environments using simple algorithm designs. One way to generate random-like scanning search is to use nonlinear dynamical systems to impart chaos into the searching robot's controller. This will result in the generation of unpredictable yet deterministic trajectories, allowing designers to control the system and achieve a high scanning coverage of an area. However, the unpredictability comes at the cost of increased coverage time and a lack of scalability, both of which have been ignored by the state-of-the-art chaotic path planners. This work introduces a new, scalable technique that helps a robot to steer away from the obstacles and cover the entire search space in a short period of time. The technique involves coupling and manipulating two chaotic systems to reduce the coverage time and enable scanning of unknown environments with different online properties. Using this new technique resulted in an average 49% boost in the robot's performance compared to the state-of-the-art planners. the overall search performance of the chaotic planner remained comparable to optimal systems while still ensuring unpredictable paths.

___________________________________________________________________

___________________________________________________________________

**Declarations**

**Funding** - The work was sponsored by the Air Force under MOU FA8750-15-3-6000. The U.S. Government is authorized to reproduce and distribute copies for Governmental purposes notwithstanding any copyright or other restrictive legends. The views and conclusions contained herein are those of the authors and should not be interpreted as necessarily representing the official policies or endorsements, either expressed or implied, of the Air Force or the U.S. Government.

**Conflicts of interest/Competing interests** - The authors declare no conflict of interest.

**Availability of data, material, or code** - https://gitlab.com/dsim-lab/online-search-of-unknown-terrains

---

[1] Corresponding author

# NOMENCLATURE

| | | | |
|---|---|---|---|
| $A$ | Arnold system's parameter | $r$ | Logistic map system's parameter |
| $B$ | Arnold system's parameter | $right$ | Right boundary |
| $C$ | Arnold system's parameter | $t$ | Time |
| $c$ | Coverage criterion factor | $tc$ | Total coverage rate |
| $c_n$ | Number of new cells visited in $[t-\Delta t: t]$ | $tc_{new}$ | Total coverage rate at $t + \Delta t$ |
| $c_{nr}$ | Total number of new and repetitive cells visited in $[t-\Delta t: t]$ | $tc_{old}$ | Total coverage rate at $t$ |
| $c_u$ | Total number of unvisited cells in $[0: t-\Delta t]$ | $t_h$ | Maximum number of attempts to avoid an obstacle |
| $coor_{obs}$ | Coordinate matrix of obstacles | $T_p$ | Matrix storing all trajectory points |
| $coor_{boundary}$ | Coordinate matrix of the map boundaries | $TP_{DS-R}$ | Temporary matrix of Arnold DS and robot coordinates |
| $c_T$ | Total number of cells | $TP_{DS-R-logistic}$ | Temporary matrix of Logistic map initial DS and robot's coordinates |
| $CT$ | Time for 90% coverage rate | $TP_{scaled-R}$ | Temporary matrix of scaled robot coordinates |
| $dc$ | Desired coverage rate | $t_{TR}$ | Time taken to travel between two zones |
| $d$ | Distance between the robot's current position and a zone's midpoint | $upper$ | Upper boundary |
| $e_p$ | Error tolerance of trajectory points | $v$ | Robot's velocity |
| $f_o$ | Obstacle/boundary offset factor | $x_n, x_{n+1}$ | Coordinates of Logistic map system at current and next iterations |
| $f$ | Scaling factor | $\widehat{X}_{n+1}, \widehat{Y}_{n+1}$ | Relative coordinates of Logistic map system in next iteration |
| $i, j$ | Incremental index | $X'_{n+1}, Y'_{n+1}$ | Mapped coordinate of Logistic map system in next iteration |
| $DS_{index}$ | Index for each DS coordinate | $X(t), Y(t)$ | Robot coordinates mapped from Arnold system |
| $left$ | Left boundary | $X_n, Y_n, X_{n+1}, Y_{n+1}$ | Robot coordinates mapped from Logistic map system in current ($n$) and next ($n+1$) iterations |
| $lower$ | Lower boundary | $X_{end}, Y_{end}$ | Zone's Midpoint coordinates |
| $l_{zone}$ | List of zones with their respective properties | $[X, Y]$ | Trajectory matrix |
| $M$ | Column coordinates | $[\widehat{X}, \widehat{Y}]$ | Relative trajectory matrix |
| $m$ | Number of logistic map trajectory points on constructed path between two zones | $[X', Y']$ | Mapped trajectory matrix |
| $N$ | Row coordinates | $x(t), y(t), z(t)$ | Coordinates of Arnold system |
| $n$ | Coordinate index number | $\Delta t$ | Time step |
| $n_{iter}$ | Number of iterations | $\Delta t_{adaptive}$ | Adaptive time step |
| $n_h$ | Number of attempts to avoid an obstacle | $\Delta t_{constant}$ | Constant time step |
| $n_{map}$ | Last row index in mapped matrix $[X', Y']$ | $\Delta n_h$ | Change in $n_h$ |
| $n_{TP_{DS-R}}$ | Last row index in $TP_{DS-R}$ | $\theta$ | Mapping variable |
| $n_{TP_{DS-R-logistic}}$ | Last row index in $TP_{DS-R-logistic}$ | $\rho$ | Zone's coverage density |

# 1 INTRODUCTION

Smart mobile robots with coverage path planning (CPP) algorithms have been increasingly employed in various applications [1-4]. Previous work using CPP algorithms have studied household environments [5], surveillance and search problems [3, 6-13], and agricultural environments [14]. CPP algorithms construct a path for the robot to visit every area within a map while simultaneously avoiding any present static obstacles. Various methods have been used in CPP algorithms such as the wavefront methods [15,, 16] and spanning trees [17,, 18], although the robot trajectories from CPP based on these methods tend to be predictable. Among existing CPP algorithms, the chaotic path planning is one of the most promising methods that are useful in surveillance and exploration applications, even when adversarial agents are present. In surveillance missions, the robot should perform an online scan of an uncertain environment without the need for the map and complete the scan in the minimum amount of time to find an object/target. This scan needs to be accomplished while simultaneously avoiding any present obstacles and adversaries in this uncertain environment. Due to their deterministic nature and the possibility of being controlled by the designer, the chaotic path planners have the potential to fulfill these requirements while remaining unpredictable to adversaries.

To generate chaotic trajectories, nonlinear chaotic dynamical systems (DS) are used to induce chaos into the robot's controller. These dynamical systems are simply a mathematical tool to generate pseudo-random trajectories that will allow the robot to theoretically span the search space without the robot having any knowledge of the search space or its current location within the search space. Nakamura and Sekiguchi [19] demonstrated that chaotic path planning is more effective than random-walk methods such as Brownian motion and Lévy flight [20]. Random-walk searches restrict the continuity in the robot's motion and often results in uneven density of coverage within the search space. In contrast, CPP algorithms are able to generate continuous robotic trajectories and examine the entire area of the map more efficiently. Various chaotic dynamical systems have used within CPP algorithms, such as the Arnold system [19, 21], Lorenz system [22], Chen system [23], and the Chua circuit [24, 25]. A major disadvantage of using chaotic dynamical systems for CPP as present in previous work is that they increase coverage time since most chaotic systems generate undispersed points across the map. Research has focused on manipulating the chaotic path planning algorithms [23, 26-28] to increase the dispersal of trajectory points across the environment map. Examples of chaos manipulation techniques for trajectory point dispersion include arccosine and arcsine transformations [28, 29] as well as random number generators [30]. The coverage rate criterion [19, 30-35] has been used by most studies to examine the efficiency of chaotic path planning algorithms for search while the coverage time was neglected. One challenge that has not been addressed by previous studies is scalability, which allows adaptation to new environments with different properties (e.g., size) and to cover them in a finite amount of time.

In this work, three chaos control methods were combined to enhance the coverage rate and to minimize coverage time in unknown environments: (1) orientation control, (2) map zoning, and (3) system scaling. Orientation control guides the robot to change direction and cover new cells in adjacent regions while map zoning guides the robot to regions that have less previously covered area. Finally, system scaling allows varying the coverage density based on theسensor range and environment size. The first two chaos control techniques were introduced in [36] where properties of the continuous Lorenz system and the discrete Hénon map system were combined to achieve fast coverage in an environment with no obstacles. However, these techniques were not as capable to adapt to environments with unknown obstacles, variations in the robot's sensing range (SR), and different environment sizes. To address these issues, a new hybrid system is introduced which includes the continuous Arnold system and the discrete Logistic map system, leading to relatively faster coverage. An obstacle avoidance technique which enhances the scalability of the path planner in environments containing unknown obstacles is also introduced. Using a combination of the hybrid system and obstacle avoidance, the robot will be able to prevent repeated coverage and achieve the desired coverage rate in unknown environments safely and efficiently.

The rest of the paper is organized as follows: Section 2 introduces the chaotic systems and how they are integrated to construct the robot's trajectory. Afterwards, Section 3 describes the path planning strategy and chaos control techniques, followed by Section 4 which discusses the obstacle/boundary avoidance methods. Then, Section 5 presents the simulation results of the robot scanning an unknown environment, both with and without obstacles, before finally Section 6 which concludes the paper and highlights the future work.

## 2 CHAOTIC DYNAMICAL SYSTEMS AND THEIR INTEGRATION IN MOBILE ROBOT

Chaotic dynamical systems are represented as continuous [19, 31, 37-39] or discrete [26-28, 40, 41] systems. The Arnold system (see Eq. (1)) [19] and the Logistic map (see Eq. (2)) [28] systems yielding minimal coverage time and are defined by the following equations:

$$dx(t)/dt = A \sin z(t) + C \cos y(t) \quad (1)$$
$$dy(t)/dt = B \sin x(t) + A \cos z(t)$$
$$dz(t)/dt = C \sin y(t) + B \cos x(t)$$

$$x_{n+1} = r x_n (1 - x_n) \quad (2)$$

where $x(t)$, $y(t)$, and $z(t)$ are the DS coordinates; $A$, $B$ and $C$ are the Arnold system parameters; $x_n$ and $x_{n+1}$ are the Logistic map system coordinate points for the current and new states respectively; and $r$ is the Logistic system parameter. To be considered chaotic, nonlinear dynamical systems should possess the following characteristics: (1) sensitivity to initial conditions (ICs) and (2) topological transitivity. In this work, we use the tuple $(x_0, y_0, z_0) = (0, 1, 0)$ as the ICs for the Arnold system and $x_0 = 0.1$ as the IC for the Logistic map system. Based on previous analysis of the parameter space, the Arnold system parameters are chosen to be $A = 0.5$, $B = 0.25$ and $C = 0.25$ which matches those in [19]. There exist other dynamical system parameters for the Arnold dynamic system that satisfy properties (1) and (2) and these parameter values will be associated with different levels of topological transitivity and changed shapes of the chaotic attractors. However, the Arnold DS parameters chosen in this work were found in the parameter space analysis to result in reasonably uniform coverage of the environment when the Arnold system is incorporated into the robot's trajectories as in the manner discussed in Section 5. For the Logistic map system, the parameter $r = 4$ is chosen to match previous literature. Fig. 1 shows the 3-D chaos attractor of the Arnold system obtained using the above-mentioned parameters. As the Logistic map is a 1-D system, its chaos attractor does not exist and is not shown.

The CPP algorithm maps the DS coordinates into the robot's kinematic equation, i.e., the DS coordinates control the robot's orientation. Fig. 2 depicts the two-wheel differential drive mobile robot used in this study. The robot comprises of two active fixed wheels and one passive caster wheel and the robot is subject to a non-holonomic constraint. Eqs. (3) and (4) represent the mapping process for the Arnold and Logistic dynamical systems, as suggested in [31] and [28], respectively:

$$dX(t)/dt = v \cos(x(t)) \tag{3}$$
$$dY(t)/dt = v \sin(x(t))$$
$$\omega(t) = d\theta(t)/dt = dx(t)/dt$$

$$\theta_n = \pi x_n \pm \frac{\pi}{2} \tag{4}$$
$$X_{n+1} = X_n + \Delta t \, v \cos \theta_n$$
$$Y_{n+1} = Y_n + \Delta t \, v \sin \theta_n$$
$$\omega_n = (\theta_{n+1} - \theta_n)/\Delta t$$

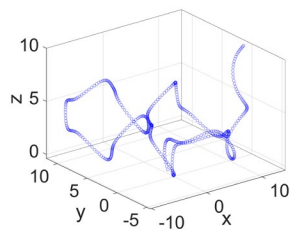

**Fig. 1.** 3-D Chaos attractor for the Arnold system.

In Eq. (3), $X(t)$ and $Y(t)$ are the robot's coordinates, $v$ is the robot's velocity, $dt$ and $\Delta t$ are time steps, and $x(t)$ is one of the Arnold system coordinates mapped into the robot's kinematic equations. In the following sections of this study, the robot coordinates are referred to as the trajectory points. It is important to note that the 3-D Arnold system has been used in this study (see Eq. (1)) and either of the other two DS coordinates, $y(t)$ or $z(t)$, can replace $x(t)$ in Eq. (3) to perform the mapping between the Arnold system and the robot's kinematic equation.

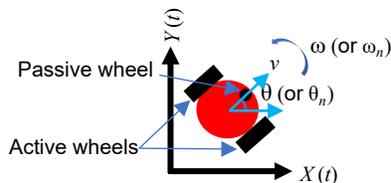

**Fig. 2.** A schematic of mobile robot's motion on a plane.

In Eq. (4), $X_n$ and $Y_n$ are the coordinates of the current states, $X_{n+1}$ and $Y_{n+1}$ are the coordinates of the new states, and $\theta_n$ is a mapping variable that maps the chaotic DS coordinates to the robot's coordinate (i.e., robot's orientation). As discrete systems usually tend to drive the robot in only one direction, the first relation in Eq. (4) is meant to prevent this issue by switching the sign next to the phase angle and directing the robot to move in another direction to effectively cover an area. For instance, if the robot is heading close to the left boundary, the algorithm will set $\theta_n$ equal to $\pi x_n - \pi/2$ and steer the robot to navigate toward the right boundary. If the opposite occurs where the robot is navigating toward the right boundary, the algorithm will set $\theta_n$ equal to $\pi x_n + \pi/2$ and steer the robot to navigate toward the left boundary. Evaluation for determining whether the sign should be positive or negative in Eq. (4) is done once per chaotic path segment. The chaotic path segments are either generated periodically by the CPP algorithm using the Logistic system or sporadically for an obstacle avoidance technique discussed in Section 3.2.1.

# 3 PATH PLANNING STRATEGY AND CHAOS CONTROL TECHNIQUES

The objective of this work is to reduce repetitive coverage and eliminate lack of scalability, both of which will lead to extended coverage time. To address these challenges, the proposed algorithm utilizes the following: (1) an orientation control technique which guides the robot to less-visited adjacent regions, (2) a map zoning technique to direct the robot to less-visited distant regions, and (3) a system scaling technique to reduce repetitive coverage and adapt to changes in the robot's sensing range and environment size. In what follows, cells refer to small segments in the map while zones indicate larger segments.

**Algorithm 1**: Chaotic path planning algorithm

**Inputs**: $A, B, C, r, TP_{DS\text{-}R}, TP_{DS\text{-}R\text{-}logistic}, f_o, f, n_h, \Delta n_h, n_{iter}, coor_{boundary}, c, dc, e_p, le, t_h, v, \Delta t_{constant}, \Delta t_{adaptive}, t$
**Outputs**: $tc, t$

1  Initialize $DS_{index}$
2  $tc \leftarrow 0$
3  $T_p \leftarrow \emptyset$
4  $tc_{new} \leftarrow 0$
5  **while** ($tc < dc$) **do**
6      $tc_{old} \leftarrow tc_{new}$
7      $\{TP_{DS-R}, TP_{scaled-R}, t, \Delta t_{adaptive}\} \leftarrow SystemScaler\,(TP_{DS-R}, f_o, f, n_{iter}, e_p, \Delta t_{adaptive}, coor_{obs}, t, A, B, C, DS_{index}, coor_{boundary}, v)$
8      $T_p \leftarrow TP_{scaled-R} \cup T_p$
9      Count the number of new and repeated cells visited in environment
10     Calculate $tc$ and $t$
11     $tc_{new} \leftarrow tc$
12     **if** ($tc < dc$) **then**
13         Stop;
14     **end**
15     **If** ($\frac{c_n}{c_{nr}} \geq c\,[\frac{c_u}{c_T}]$) **then**
16         **if** ($tc_{new} == tc_{old}$) **then**
17             $DS_{index} \leftarrow DS_{index}+1$
18         **end**
19         $TP_{scaled\text{-}R}(1,:) \leftarrow TP_{scaled\text{-}R}(n_{TP_{DS\text{-}R}},:)$
20         $TP_{DS\text{-}R}(1,:) \leftarrow TP_{DS\text{-}R}(n_{TP_{DS\text{-}R}},:)$
21         Total unvisited cells $\leftarrow$ Total unvisited cells — new cells visited
22     **else**
23         $\{TP_{DS-R}, TP_{scaled-R}, t, T_p, tc\} \leftarrow MapZoning\,(TP_{DS\text{-}R}, TP_{DS\text{-}R\text{-}logistic}, TP_{scaled\text{-}R}, f_o, f, T_p, coor_{obs}, \Delta t_{constant}, t, r, coor_{boundary}, \Delta n_h, t_h, dc, tc, v)$
24     **end**
25 **end**

The following simplifying assumptions to reduce the computational complexity of the simulation task: (1) the environments are square rooms with boundaries on all sides; (2) the robot's velocity is a constant 1 m/s to simulate a small two-wheeled differential drive robot; (3) there are only three different relatively large world sizes (50 m × 50 m, 100 m × 100 m, 200 m × 200 m) examined; (4) the sensing range determines the cell dimensions in each environment, e.g., cell dimensions of 4 m × 4 m correspond to four meters sensing range; (5) all systems are run to achieve the desired coverage rate of 90%;

(6) in simulation, the robot uses the coordinate matrix of known, stationary obstacles ($coor_{obs}$) and map boundaries ($coor_{boundary}$) to detect obstacles' and walls' boundaries, respectively, instead of utilizing sensors in real life; and (7) the robot can start the coverage task from any point in the environment, as long as the selected point is not occupied by an obstacle. In this study, the selected starting point for a robot scanning an environment with either zero obstacles or one obstacle was $X(t) = 0.5$ m and $Y(t) = 0.5$ m. For the environment with four and five obstacles, the robot's starting points at the beginning of the coverage task were as follows: $X(t) = 15$ m and $Y(t) = 5$ m in a 50 m × 50 m environment; $X(t) = 30$ m, $Y(t) = 10$ m in a 100 m × 100 m environment; and $X(t) = 60$ m, $Y(t) = 20$ m in a 200 m × 200 m environment. While these assumptions dramatically reduce the computational complexity of the simulation task, they may not be valid for real-world industry operations. However, the results presented later in Section 5 are still applicable to industry operations and can be used for estimating the robot performance in other operational environments. Assumptions (1) and (6) are consistent with an occupancy grid commonly used for robot exploration. Occupancy grid can also be updated with discovered obstacles if assumption (6) is not valid. The coverage time of other robotic platforms can be estimated based on scaling to the robot velocity and sensing range values used in assumptions (2) and (4) of this study as the coverage time is roughly inversely proportional to both velocity and sensing range. The coverage time for other sized environments can also be estimated based on assumption (3) as the coverage time tends to grow proportionally to the area of the desired coverage space.

Algorithm 1 uses the Arnold system as the primary DS to scan the environment continuously until the desired coverage ($dc$) is achieved. Each iteration invokes the *SystemScaler* function (Algorithm 3) to generate trajectory points for the robot using the Arnold system to scan the area and add the generated points (stored under temporary matrix $TP_{scaled-R}$) to the storage matrix $T_p$. The chaotic system scaling is determined by the scaling factor ($f$), which indicates to the *SystemScaler* to scale the trajectory points before returning them. After updating the trajectory points, the chaotic path planning algorithm updates the coverage rate ($tc$) and time ($t$), respectively.

To determine whether to focus on orientation or zoning, we use the following criterion:

$$\frac{c_n}{c_{nr}} \geq c \left[\frac{c_u}{c_T}\right] \tag{5}$$

where $c_n$ and $c_{nr}$ are respectively the number of new cells and total number of new and repetitive cells visited in the time interval [$t—\Delta t$: $t$]; $c_u$ is the total number of unvisited cells in the time interval [$0$: $t—\Delta t$]; and $c_T$ is the total number of cells. Eq. (5) indicates whether the current trajectory has recently been successful in visiting new cells in the current zone. If the trajectory was unsuccessful, the robot should leave the current trajectory and zone and travel to another zone where it starts a new trajectory. In this case, Algorithm 1 calls the function *MapZoning* (Algorithm 2) to find the next best zone for coverage as well as generating a direct chaotic path between the current robot position and the new zone midpoint using the Logistic map system. Algorithm 2 utilizes Algorithm 4 (*LogisticObstacleAvoid*) to ensure the direct chaotic path returned from Algorithm 2 is an obstacle free path to the new zone. Once the robot arrives at the new zone, Algorithm 3 (*SystemScaler*) will generate $n$ new trajectory points for the robot to continue the coverage task in the new zone.

## 3.1 Orientation Control

Each Arnold system coordinate is associated with an index number ($DS_{index}$), with index numbers 1, 2, and 3 representing coordinates $x(t)$, $y(t)$, and $z(t)$ respectively. At the start of the coverage task, an initial index number ($DS_{index}$) is selected to be mapped into the robotic kinematic equations. To avoid repetitive coverage, Algorithm 1 keeps track of the coverage rate in the previous iteration ($tc_{old}$) and the current iteration ($tc_{new}$). During the coverage task, Algorithm 1 records the coverage rate obtained in the previous iteration using variable $tc_{old}$ and sets the $tc$ calculated in the current iteration to $tc_{new}$. If there is no change identified in the coverage rate between iterations, i.e. $tc_{new}$ is equal to $tc_{old}$, the robot moves to cover new cells in neighboring regions by switching $DS_{index}$ to another index. Algorithm 1 will then define the robot and DS initial conditions by replacing the first entries of the temporary matrixes $TP_{scaled-R}$ and $TP_{DS-R}$ by their current last entries as denoted by the row index, $n_{TP_{DS-R}}$. The matrixes $TP_{scaled-R}$ and $TP_{DS-R}$ respectively contain the scaled and unscaled robot's coordinates of the Arnold system generated by the *SystemScaler* function. The algorithm then updates the number of unvisited cells in order to calculate the criterion relation in the next iteration. The performance can be further improved using a second technique called map zoning which is discussed in Section 3.2.

## 3.2 Map Zoning

To further scan for new cells and increase dispersion of the robot's path, the robot uses a map zoning method that breaks the map down into 16 individual zones and generates a priority list ($l_{zone}$) containing: each zone's midpoint, the distance ($d$) between the robot's current position and the midpoint of each zone, the robot's current zone and position, and the coverage density ($\rho$) of each zone. For a robot traversing an obstacle-free environment, the algorithm prioritizes density over distance for selecting zones for coverage. For environment with obstacles, distance is prioritized over density to select the new zone for coverage. In the obstacle scenario, the algorithm will construct a collision-free path to direct the robot from its current position to the new zone midpoint to begin a chaotic search of the new zone. If the robot cannot reach the midpoint of the new zone within a certain number of attempts ($t_h$), it will perceive that zone as being inaccessible at the time.

### 3.2.1 *Logistic Map Trajectory*

In the map zoning technique, the robot uses the Logistic map system to travel chaotically from its current position to the new zone's midpoint. To ensure that the robot navigates to the specified zone, Eq. (6) maps the Logistic map robot's trajectory (obtained from Eq. (4)) onto a direct path connecting the current position and the new starting point and verifies if there is a safe distance between the mapped points and the environment boundaries and obstacles. Each time Algorithm 1 calls the *MapZoning* function, it passes the matrix $TP_{DS-R-logistic}$ which initially only includes the DS initial conditions and later becomes populated with new entries of the DS and robot coordinates. The robot's initial coordinates on the path between the two zones is the last position coordinates of the robot in the current zone (i.e., last point of the previous Arnold trajectory).

To determine the number of trajectory points required to travel between the current robot position and the new zone's midpoint, the *MapZoning* function constructs a time vector ($t_{TR}$) with a constant step size of 0.1s ($\Delta t_{constant}$) and a length $m$. The vector starts from 0 and ends at $d_{min}/v$. Knowing the length of this time vector, Algorithm 2 uses Eqs.

(2) and (4) to generate *m* new DS and robot position coordinates and assigns those to matrix $TP_{DS-R-logistic}$; the generated robot trajectory points are referred here as matrix ([*X*, *Y*]). If any of the Logistic map trajectory points approach infinity or become not a number (NaN), Algorithm 2 terminates with an error. This might happen due to choosing non-chaotic DS initial conditions for the Logistic map system at the beginning of the simulation. Otherwise, Algorithm 2 generates *m* evenly spaced relative trajectory points to map the

---

**Algorithm 2**: *MapZoning* ($TP_{DS-R}$, $TP_{DS-R-logistic}$, $TP_{scaled-R}$, $f_o$, $f$, $T_p$, $coor_{obs}$, $\Delta t_{constant}$, $t$, $r$, $coor_{boundary}$, $n_h$, $\Delta n_h$, $t_h$, $dc$, $tc$, $v$)

---

    **Inputs:** $TP_{DS-R}$, $TP_{DS-R-logistic}$, $TP_{scaled-R}$, $f_o$, $f$, $T_p$, $coor_{obs}$, $\Delta t_{constant}$, $t$, $r$, $coor_{boundary}$, $n_h$, $\Delta n_h$, $t_h$, $dc$, $tc$, $v$
    **Outputs:** $TP_{DS-R}$, $TP_{scaled-R}$, $t$, $T_p$, $tc$
1    $TP_{DS-R-logistic}(1, 2:3) \leftarrow TP_{scaled-R}(n_{TP_{DS-R}}, :)$
2    **while** ($tc < dc$) **do**
3        Generate the list $l_{zone}$
4        **while** ($l_{zone} \neq \emptyset$) **do**
5            Find the zone in $l_{zone}$ associated with least distance $d_{min}$
6            **if** (multiple zones associate with $d_{min}$) **then**
7                $[X_{end}, Y_{end}] \leftarrow$ midpoint coordinates associated with $\rho_{min}$ and $d_{min}$
8            **else**
9                $[X_{end}, Y_{end}] \leftarrow$ midpoint coordinates associated with $d_{min}$
10           **end**
11           $t_{TR} \leftarrow [0 : \Delta t_{constant} : d_{min}/v]$
12           $TP_{DS-R-logistic} \leftarrow$ Eqs. (2) and (4)
13           **if** ($TP_{DS-R-logistic}(:, 2:3) == \infty$ or NaN) **then**
14                Stop;
15           **end**
16           Generate matrix $[\hat{X}, \hat{Y}]$
17           **for** ($i = 1, 2, …, m$) **do**
18                $[X'(i,1), Y'(i,1)] \leftarrow$ Eq. (6)
19                **if** ($[X'(i, 1), Y'(i, 1)]$ is outside or close to a boundary) **then**
20                    $[X'(i, 1), Y'(i, 1)] \leftarrow$ Eq. (7)
21                **end**
22                **if** ($[X'(i, 1), Y'(i, 1)]$ is outside a boundary) **then**
23                    Stop;
24                **end**
25           **end**
26           $\{[X', Y'], t, n_h\} \leftarrow LogisticObstacleAvoid ([X', Y'], [\hat{X}, \hat{Y}], TP_{DS-R-logistic}, [X_{end}, Y_{end}], t, \Delta t_{constant}, coor_{obs}, f_o, t_h, m)$
27           $T_p \leftarrow [X', Y'] \cup T_p$
28           Count the number of new and repeated cells visited
29           Calculate $tc$ and $t$
30           $TP_{scaled-R}(1, :) \leftarrow [X'(n_{map},1), Y'(n_{map},1)]$
31           $TP_{DS-R}(1, :) \leftarrow [TP_{DS-R}(n_{TP_{DS-R}}, 1:3), X'(n_{map},1) / f, Y'(n_{map},1) / f]$
32           **if** ($[X'(n_{map},1), Y'(n_{map},1)] == [X_{end}, Y_{end}]$ or $tc > dc$) **then**
33                Break;
34           **end**
35           **if** ($n_h == t_h$) **then**
36                $TP_{DS-R-logistic}(2: n_{TP_{DS-R-logistic}}, :) \leftarrow \emptyset$
37                $TP_{DS-R-logistic}(1, 2:3) \leftarrow [X'(n_{map},1), Y'(n_{map},1)]$
38                Eliminate the zone that the robot failed to reach from $l_{zone}$
39                **if** ($l_{zone} == \emptyset$) **then**
40                    $t_h \leftarrow t_h + \Delta n_h$
41                **end**
42           **end**
43        **end**
44        **if** ($[X'(n_{map},1), Y'(n_{map},1)] == [X_{end}, Y_{end}]$ or $tc > dc$) **then**
45           Break;
46        **end**
47    **end**

robot's trajectory points onto a direct chaotic path between the starting and final trajectory points. The start point of the relative trajectory matrix is the difference between the first point of the Logistic map trajectory and the robot's current position, and the end point is the difference between the last point of Logistic map trajectory and the midpoint of the new zone (i.e., [$X_{end}$, $Y_{end}$]). Using the Logistic map ([$X$, $Y$]) and the relative trajectory points ([$\hat{X}$, $\hat{Y}$]), Algorithm 2 generates $m$ mapped trajectory points and uses Eq. (6) to generate the mapped trajectory points ([$X'$, $Y'$]) with the first point being the robot's current position and the last point being the midpoint of the new zone:

$$X' = X + \hat{X} \qquad (6)$$
$$Y' = Y + \hat{Y}$$

If any points are located outside the environment boundaries or are closer than a certain tolerance to any boundary, Algorithm 2 uses Eq. (7) and the boundary avoidance method discussed in Section 4 to translate the mapped trajectory point to a safe distance from the boundary to avoid collision. Improper choices of the DS initial conditions for the Logistic map system may generate mapped points far away from the boundary, in which case the simulation will terminate. Otherwise, the *LogisticObstacleAvoid* function (Algorithm 4) will construct a new path from the previous valid mapped point to avoid the obstacle.

In the event that the robot reaches the midpoint of the new zone, it should be prepared to start coverage of the new zone using the Arnold system in the next iteration. Therefore, Algorithm 2 assigns the last mapped trajectory points in matrix [$X'$, $Y'$] as the robot's ICs to the first entry of matrix $TP_{scaled-R}$, and assigns the last DS coordinates of the Arnold system and unscaled mapped points to the first entry of matrix $TP_{DS-R}$. Otherwise, if the algorithm determines that the robot has reached the maximum number of attempts to avoid the obstacle, it will remove all the entries of matrix $TP_{DS-R-logistic}$ except the Logistic map DS initial conditions and define the last mapped trajectory point in matrix [$X'$, $Y'$] as the robot's current position The *MapZoning* algorithm will attempt to traverse to all other zone midpoints, from zone with the least coverage to the zone with the highest coverage, successively until the robot has successfully reached a zone midpoint. In the event that no zone midpoints are reachable, then the algorithm will terminate and the robot will again traverse according to Algorithm 1. Within cluttered environments, *MapZoning* allows for the robot to attempt to navigate to reachable locations that will best improve the coverage rate although there may be no reachable locations. If there are no reachable locations, the robot will return to searching with chaotic paths generated by the Arnold dynamical system. This allows for the robot the chance to navigate within the environment so that when *MapZoning* is called again, the robot will be able to reach a different zone midpoint.

### 3.3 System Scaling

In the system scaling technique, $f$ is applied to scale the robot's trajectories, allowing it to adjust the coverage density based on the robot's sensing range and to vary the coverage extent based on the environment size. To determine the influence of the scaling technique on the robot's trajectory, the Arnold system with three different scaling factors (0.1, 1 and 2) were used to cover the same environment in the same time period of $2.16 \times 10^3$ seconds. The robot has a sensing range of 1 m in this example. Fig. 3 depicts the results which show the strong influence of scaling factor on the trajectories extent and coverage density. As observed in Figs. 3 (a)-(c), the robot manages to cover 18%, 40%, and 54% of the map with the scaling factor of 0.1, 1, and 2, respectively. In this specific example, the increase in the scaling factor, and therefore the coverage extent, has sped up the coverage task. Section 5 describes the results associated with additional system scaling tests.

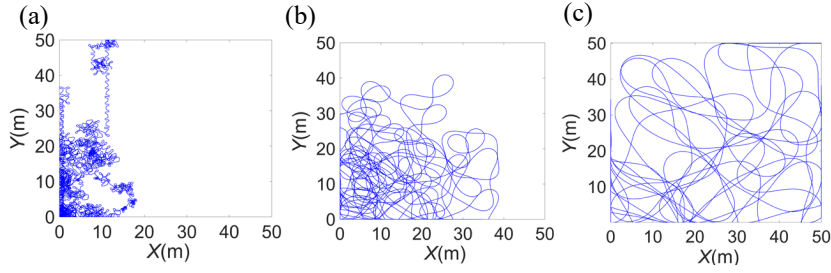

**Fig. 3.** Robot's trajectory created using the Arnold system in obstacle-free environment (0% occupancy) with: (a) $f = 0.1$, (b) $f = 1$, (c) $f = 2$.

Algorithm 3 generates and returns the robot's trajectory using the Arnold system. First, the algorithm scales the robot's current position coordinates in matrix $TP_{DS-R}$ using $f$ and assigns the results to the first row of $TP_{scaled-R}$. The algorithm will then use Eqs. (1) and (3) to generate $n$ new trajectory points. These points create the proposed path for the robot to scan the current region. The algorithm uses $n = 1000$ based on analysis of initial simulation results. Algorithm 3 uses a Runge-Kutta fourth order (RK4) method to propagate Eqs. (1) and (3) forward in time as suggested in [31]. However, instead of a constant time step, the present study proposes to use an adaptive time step ($\Delta t_{adaptive}$). The preliminary study demonstrated that a constant time step causes error in calculating the robot trajectory points that accumulates over time. The study also showed strong influence of the initial choice of adaptive time step on accuracy of the points and coverage time. The optimal value for initial time step was found to be 0.1 seconds.

---

**Algorithm 3:** *SystemScaler* ($TP_{DS-R}, f_o, f, n_{iter}, e_p, \Delta t_{adaptive}, coor_{obs}, t, A, B, C, DS_{index}, coor_{boundary}, v$)

**Inputs:** $TP_{DS-R}, f_o, f, n_{iter}, e_p, \Delta t_{adaptive}, coor_{obs}, t, A, B, C, DS_{index}, coor_{boundary}, v$
**Outputs:** $TP_{DS-R}, TP_{scaled-R}, t, \Delta t_{adaptive}$

1  $TP_{scaled-R}(1, 1{:}2) = f \; TP_{DS-R}(1, 4{:}5)$
2  **for** ($i = 1, 2, \ldots, n_{iter}$,) **do**
3    $[x(t), y(t), z(t), X(t), Y(t)] \leftarrow$ Eqs. (1) and (3)
4    $[x_{half}(t), y_{half}(t), z_{half}(t), X_{half}(t), Y_{half}(t)] \leftarrow$ Eqs. (1) and (3)
5    **if** ($|(X(t) - X_{half}(t)| > e_p$ or $|(Y(t) - Y_{half}(t)| > e_p$) **then**
6      $TP_{DS-R}(i+1, :) \leftarrow [x_{half}(t), y_{half}(t), z_{half}(t), X_{half}(t), Y_{half}(t)]$
7      $\Delta t_{adaptive} \leftarrow \Delta t_{adaptive}/2$
8    **else**
9      $TP_{DS-R}(i+1, :) \leftarrow [x(t), y(t), z(t), X(t), Y(t)]$
10   **end**
11   **if** ($TP_{DS-R}(i+1, :) == \infty$ or NaN) **then**
12     Stop;
13   **end**
14   **if** ($TP_{DS-R}(i+1, 4{:}5)$ is outside or close to a boundary) **then**
15     $TP_{DS-R}(i+1, 4{:}5) \leftarrow$ Eq. (7)
16   **end**
17   **if** ($TP_{DS-R}(i+1, 4{:}5)$ is inside or close to an obstacle) **then**
18     $TP_{DS-R}(i+1, 4{:}5) \leftarrow$ Eq. (8)
19   **end**
20   $TP_{scaled-R}(i+1, 1{:}2) = f \; TP_{DS-R}(i+1, 4{:}5)$
21   **if** ($TP_{scaled-R}(i+1, 1{:}2)$ is outside a boundary) **then**
22     Stop;
23   **end**
24   $t \leftarrow t + \Delta t_{adaptive} \; f$
25 **end**

To apply the adaptive RK4 method, we compare the difference of the robot coordinates obtained using different time steps, assigning the dynamic and robot trajectory points associated with $\Delta t_{adaptive}$ to $TP_{DS\text{-}R}$ if the output does not exceed the error tolerance ($e_p$). The simulation will terminate if any nonfinite trajectory points are observed among the newly generated points, similarly to Algorithm 2. Algorithm 3 transfers any trajectory point that lie outside or close to any boundary or obstacle using Eqs. (7) and (8). All the generated robot trajectory points are then scaled using $f$, with the results stored in $TP_{scaled\text{-}R}$. Finally, when updating the current time, $\Delta t_{adaptive}$ is multiplied by $f$ to maintain a constant robot velocity.

## 4 OBSTACLE AVOIDANCE TECHNIQUES
### 4.1 Boundary Avoidance

Upon integrating either the Arnold or Logistic map systems and incorporating either Eq. (3) or (4) for mapping, a certain number of the robot's coordinates either lie very close to the map boundaries or outside them. Eq. (7) adapts a mirror-mapping technique as suggested in [31] to translate the robot's coordinates that are close to the boundaries or outside the map using coordinate transformations:

$$X = -X + 2(M_{left} \pm f_o) \qquad (7)$$
$$Y = -Y + 2(N_{upper} \pm f_o)$$

where $M$ and $N$ respectively represent the column and row coordinates; *left* and *upper* indicate the left and upper boundaries; and $f_o$ is a factor that determines how far the mirrored points offset into the map with respect to the boundaries. The value of $f_o$ is a chosen design parameter based on the robot's size and sensor characteristics. This study considers $f_0 = 0.5\ m$ as a safe distance from the boundaries.

Eq. (7) changes its variables based on the trajectory point's location with respect to the boundaries. For instance, if the $X$- coordinate of the point was positioned outside or close to the left boundary, then Eq. (7) will shift only the $X$-component of the point and the $Y$-component will remain unchanged. If the $X$-coordinate was positioned outside or close to the right boundary, then $M_{left}$ is substituted by $M_{right}$ to translate the $X$-component. A similar procedure would take place if the $Y$-component lay outside or close to the boundaries. If both coordinates lay outside or near a boundary, then both components would be translated, depending on the position of the coordinates with respect to the boundaries. The sign before $f_o$ will change as well depending on the trajectory point's location with respect to the boundaries. A '—' sign before $f_o$ will appear when the points lay outside or close to the right/upper boundaries. A '+' sign will take place when the points lay outside or close to the left/lower boundaries. By incorporating this technique of mirror mapping, the robot was confined within the boundaries for its coverage task. Section 4.2 establishes techniques to avoid obstacles throughout the environment.

### 4.2 Obstacle Avoidance

The robot uses different obstacle avoidance techniques depending on whether the Arnold or the Logistic map system is employed for navigation as elaborated in Sections 4.2.1 and 4.2.2. Both obstacle avoidance techniques discussed in the following sections assume that obstacles present in the environment are static and the obstacles are represented in a local occupancy grid. These obstacle avoidance methods can be extended to a moving obstacle scenario but safety guarantees involving these extensions are not discussed in this work. The moving obstacle cases introduces numerous difficulties and their discussion and analysis are beyond the scope of this work.

### 4.2.1 Obstacle Avoidance for Arnold System

The obstacle avoidance designed for the Arnold system enables the robot to sense the boundaries of all known obstacles and determines if the generated trajectory point needs to be altered to avoid any of the obstacles. Eq. (7) was presented in a previous section for boundary avoidance but can also be used for obstacle avoidance. However, Eq. (7) might place the mirror-mapped points in a manner that incurs a harsh trajectory change such that a trajectory controller may exceed a pre-determined tracking error limit. As a result, the robot would have to increase its linear and angular velocity to travel from the current position to the mirror-mapped point and hence violate the constant robot velocity assumption made in this study. Therefore, this study established Eq. (8) to position the robot away from the obstacle using $f_o$, while maintaining a constant velocity for the robot:

$$X = M_{left} \pm f_o$$
$$Y = N_{upper} \pm f_o \quad (8)$$

Eq. (8) resembles Eq. (7) with the same variables. Similar to Eq. (7), the row and column coordinates and sign before $f_o$ change depending on where the trajectory point is located with respect to the obstacle. When using trajectories generated by the Arnold system, the above-mentioned technique enabled the robot to avoid the obstacles impeccably given an appropriate Runge-Kutta time step is selected. Since all points are generated discretely using a Runge-Kutta method, all points have a maximum travel distance from the previous point which is the vehicle velocity times the Runge-Kutta time step. If this maximum travel distance is less than half the resolution of the occupancy grid, then Eq. (8) can at most map the invalid point to a diagonally adjacent occupancy cell with the mapped point being at most $\sqrt{2}$ times the maximum travel distance away from the previous point. Hence, care should be taken to ensure a sufficiently small Runge-Kutta timestep is selected such that the worst-case mapping is within the tolerated error bounds of the tracking controller. In most point mapping cases, the collision avoidance maps an infeasible point closer to the previous point which reduces the workload of the trajectory tracking controller. While the Logistic map system uses Eq. (8) for mapping as well, it is not as straightforward to avoid obstacles while still navigating towards a specific goal. Section 4.2.2 develops a new obstacle avoidance technique for Logistic map-based navigation.

### 4.2.2 Obstacle Avoidance for Logistic map system

In an environment with unknown obstacles, the robot might face two challenges when using the Logistic map system to travel to new the zone midpoint. Either the midpoint might be located in a zone surrounded by an unknown obstacle or the path generated by the Logistic map system might pass through an obstacle. Without a rigorous obstacle avoidance technique in place, the robot might spend considerable time on potentially finding a way to reach the target midpoint. To prevent this issue, a new method encompassed within Algorithm 4 was established to avoid obstacles when traveling between two points in online coverage problems.

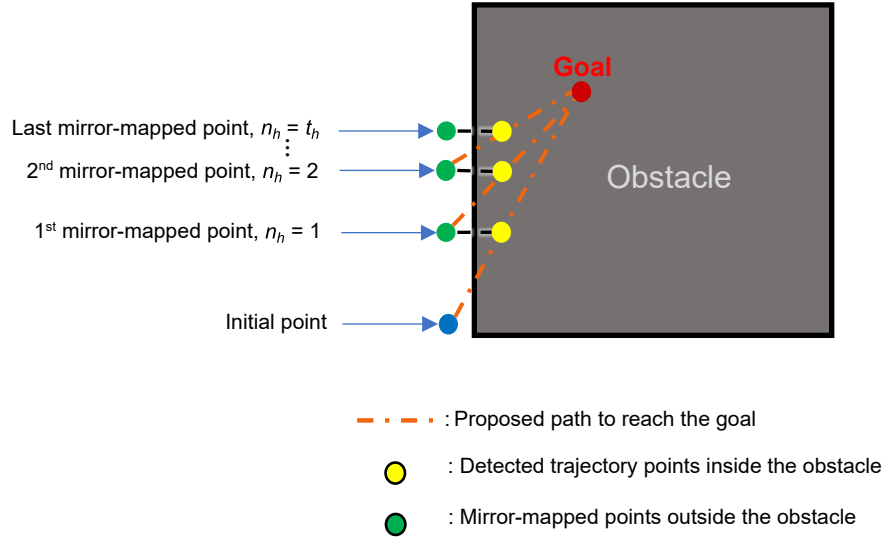

**Fig. 4.** A schematic of the obstacle avoidance technique using Logistic map.

Fig. 4 illustrates the technique used to find such a path. In this figure, the midpoint of the new zone (goal) is located inside an obstacle. At the beginning, the algorithm constructs a path to reach the goal from the robot's initial point or its current position. The robot

---

**Algorithm 4**: *LogisticObstacleAvoid* ([$X'$, $Y'$], [$\hat{X}$, $\hat{Y}$], $TP_{DS\text{-}R\text{-}logistic}$, [$X_{end}$, $Y_{end}$], $t$, $\Delta t_{constant}$, $coor_{obs}$, $f_o$, $t_h$, $m$)

**Inputs:** [$X'$, $Y'$], [$\hat{X}$, $\hat{Y}$], $TP_{DS\text{-}R\text{-}logistic}$, [$X_{end}$, $Y_{end}$], $t$, $\Delta t_{constant}$, $coor_{obs}$, $f_o$, $t_h$, $m$
**Outputs:** [$X'$, $Y'$], $t$, $n_h$

1  **for** ($i = 1, 2, …, m$) **do**
2    **if** ([$X'(i,1)$, $Y'(i,1)$] is inside an obstacle) **then**
3      [$X'(i+1{:}m, 1)$, $Y'(i+1{:}m, 1)$] ← ∅
4      [$\hat{X},\hat{Y}$] ← ∅
5      [$X'(i, 1)$, $Y'(i, 1)$] ← Eq. (8)
6      $m \leftarrow m - i + 1$
7      Generate matrix [$\hat{X}, \hat{Y}$] using $TP_{DS\text{-}R\text{-}logistic}(i{:}\,n_{TP_{DS\text{-}R\text{-}logistic}}, 2{:}3)$
8      $n_h \leftarrow 1$
9      **while** ($n_h < t_h$) **do**
10       **for** ($i = i, i+1, …, n_{TP_{DS\text{-}R\text{-}logistic}}$) **do**
11         [$X'(i,1)$, $Y'(i,1)$] ← Eq. (6)
12         **if** ([$X'(i,1)$, $Y'(i,1)$] is inside an obstacle) **then**
13           [$\hat{X},\hat{Y}$] ← ∅
14           [$X'(i, 1)$, $Y'(i, 1)$] ← Eq. (8)
15           $m \leftarrow m - i + 1$
16           Generate matrix [$\hat{X}, \hat{Y}$] using $TP_{DS\text{-}R\text{-}logistic}(i{:}\,n_{TP_{DS\text{-}R\text{-}logistic}}, 2{:}3)$
17           $n_h \leftarrow n_h + 1$
18           Break
19         **else**
20           $t \leftarrow t + \Delta t_{constant}$
21         **end**
22       **end**
23       **if** ($i == n_{TP_{DS\text{-}R\text{-}logistic}}$) **then**
24         Break
25       **end**
26     **end**
27     Break
28   **else**
29     $t \leftarrow t + \Delta t_{constant}$
30   **end**
31 **end**

travels on this path until it detects the obstacle and realizes that the next point is inside the obstacle. The algorithm uses the mirror-mapping technique (Eq. (8)) to transfer the detected point away from the obstacle. The algorithm considers this as the first attempt ($n_h = 1$) to travel to the goal. The algorithm will then construct a new path from the mirror-mapped point to the goal. The variable $n_h$ increments each time the attempt to reach the goal is not successful, i.e., the constructed path passes through the obstacle. The procedure to find an obstacle free path is repeated until the robot reaches the goal or $n_h$ reaches the maximum number of attempts ($t_h$). In the latter case, the robot realizes that it is either impossible or very time-consuming to travel to the target zone and therefore finds the next best zone to travel to by using Algorithm 2 as discussed in section 3.2.1.

For a 50 m ×50 m environment, $t_h$ is initially set to 10 at the start of the coverage task. For environment sizes of 100 m × 100 m and 200 m × 200 m, a higher initial $t_h$ is considered. As the obstacles size increases in larger environments, the robot should make more attempts to avoid them. These initial $t_h$ values allowed the robot to reach the desired zone with reasonable efforts and at the same time prevented an increase in the coverage time. With the chaos control and obstacle avoidance techniques induced into the chaotic path planning algorithm, the robot was able to visit all regions in the map, avoiding both boundaries and obstacles, to cover 90% of the map in a short time as demonstrated in Section 5.

## 5 RESULTS

This section compares the performance of systems with different chaos control methods against the original Arnold system in unknown environments with different obstacle configurations and sizes. All tests were performed on laptop with an Intel® i7-10750 CPU. Additionally, a sample codebase used for analyzing the system performances is available for public use[2]. Table 1 compares the performance of the Arnold system before and after employing a single or a combination of chaos control techniques, while varying the environment size, robot SR, and number of obstacles ($N_{obs}$). The values of $c$, the initial $DS_{index}$, and $f$ were selected to achieve the minimum coverage time for each chaotic path planning algorithm. The initial value selected for $DS_{index}$ is critical as it can significantly change the path planner's performance and influence the coverage time. To determine the optimal values of $c$, initial $DS_{index}$, and $f$ for each of the cases presented in Table 1, a series of trial and error tests were conducted. In these tests, the parameter $c$ was varied between 0.1 and 1, the initial $DS_{index}$ was selected from the valid indices, and $f$ was chosen in the range from 1 to 6. The $c$ values below 0.1 were not considered in these tests as those values would not provide enough time for the robot to properly explore the current zone. For the factor $f$, the initial tests indicated that the values over 6 or under 1 are not effective and result in a significant increase in the coverage time for the examined environments. Therefore, those values were abandoned in the future tests. The conducted tests provided the best values for parameters $c$, $DS_{index}$, and $f$ for each method and environment; Table 1 uses these values to compare the performance of different methods. Table 1 also incorporates the results of coverage using the Logistic map system (case 11) to demonstrate the Arnold system's superior performance in covering the areas. For comparisons to other work, Table 1 compares the coverage time of the Arnold-based systems with that of an optimal path planner ($T_{Opt}$) using the performance ratio $PR$ ($= CT/T_{Opt}$). The optimal path planner uses back and forth motions to cover the space (see Fig. 5). Sections 5.1 and 5.2 discuss the results in small environments (50m×50m) and large environments (100m×100m and 200m×200m) respectively.

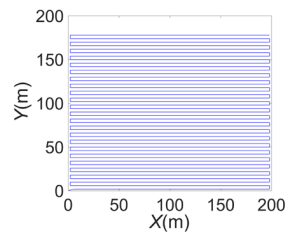

---

[2] https://gitlab.com/dsim-lab/online-search-of-unknown-terrains

**Fig. 5.** Robot's optimal trajectories (SR=4 m) for 90% coverage of the space in an obstacle-free environment (0% occupancy).

Table 1: Performance of different systems in various unknown environments.

| System type | No. | Space size | SR (m) | $f$ | $N_{obs}$ | $c$ | Initial $DS_{index}$ | $CT$ | % (hours) improv. w.r.t original Arnold | $PR$ |
|---|---|---|---|---|---|---|---|---|---|---|
| Original Arnold | 1 | 50m × 50m | 1 | N/A | 0 | N/A | 3 | $6.95 \times 10^3$ | N/A | 3.1 |
| | 2 | 50m × 50m | 1 | N/A | 1 | N/A | 3 | $6.60 \times 10^3$ | N/A | 3.5 |
| | 3 | 50m × 50m | 1 | N/A | 4 | N/A | 3 | $6.05 \times 10^3$ | N/A | 3.5 |
| | 4 | 50m × 50m | 1 | N/A | 5 | N/A | 2 | $1.04 \times 10^4$ | N/A | 7.9 |
| | 5 | 100m × 100m | 1 | N/A | 0 | N/A | 3 | $3.24 \times 10^4$ | N/A | 3.6 |
| | 6 | 100m × 100m | 1 | N/A | 5 | N/A | 3 | $2.77 \times 10^4$ | N/A | 5.1 |
| | 7 | 200m × 200m | 1 | N/A | 0 | N/A | 3 | $1.72 \times 10^5$ | N/A | 6.5 |
| | 8 | 200m × 200m | 1 | N/A | 5 | N/A | 3 | $9.92 \times 10^4$ | N/A | 4.7 |
| | 9 | 200m × 200m | 4 | N/A | 0 | N/A | 3 | $5.36 \times 10^4$ | N/A | 6.0 |
| | 10 | 200m × 200m | 4 | N/A | 5 | N/A | 3 | $3.51 \times 10^4$ | N/A | 6.6 |
| Logistic map | 11 | 50m × 50m | 1 | N/A | 0 | N/A | N/A | $2.34 \times 10^4$ | -236.7 (~ 4.57 hrs↑) | 10 |
| Arnold with Index Switching | 12 | 50m × 50m | 1 | N/A | 0 | N/A | 3 | $6.40 \times 10^3$ | 7.9 (~ 0.15 hrs↓) | 2.8 |
| | 13 | 50m × 50m | 1 | N/A | 1 | N/A | 3 | $5.45 \times 10^3$ | 17.4 (~ 0.32 hrs↓) | 2.9 |
| | 14 | 50m × 50m | 1 | N/A | 4 | N/A | 2 | $6.50 \times 10^3$ | -7.4 (~ 0.13 hrs↑) | 3.8 |
| | 15 | 50m × 50m | 1 | N/A | 5 | N/A | 3 | $7.40 \times 10^3$ | 28.8 (~ 0.83 hrs↓) | 5.6 |
| | 16 | 100m × 100m | 1 | N/A | 0 | N/A | 3 | $3.89 \times 10^4$ | -20.1 (~ 1.81 hrs↑) | 4.2 |
| | 17 | 100m × 100m | 1 | N/A | 5 | N/A | 3 | $2.13 \times 10^4$ | 23.1 (~ 1.78 hrs↓) | 4.0 |
| | 18 | 200m × 200m | 1 | N/A | 0 | N/A | 3 | $1.48 \times 10^5$ | 14.0 (~ 6.67 hrs↓) | 4.1 |
| | 19 | 200m × 200m | 1 | N/A | 5 | N/A | 3 | $8.35 \times 10^4$ | 15.8 (~ 4.36 hrs↓) | 3.9 |
| Arnold with Map Zoning | 20 | 200m × 200m | 1 | N/A | 0 | 0.25 | 3 | $8.62 \times 10^4$ | 49.9 (~ 23.8 hrs↓) | 2.4 |
| | 21 | 200m × 200m | 1 | N/A | 5 | 0.15 | 3 | $7.78 \times 10^4$ | 21.6 (~ 5.94 hrs↓) | 3.7 |
| | 22 | 200m × 200m | 4 | N/A | 0 | 0.20 | 3 | $3.57 \times 10^4$ | 33.4 (~ 4.97 hrs↓) | 4.0 |
| | 23 | 200m × 200m | 4 | N/A | 5 | 0.15 | 3 | $2.21 \times 10^4$ | 37.0 (~ 3.61 hrs↓) | 4.2 |
| Arnold with Scaling | 24 | 200m × 200m | 1 | 1.90 | 0 | N/A | 3 | $1.13 \times 10^5$ | 34.3 (~16.39 hrs↓) | 3.1 |
| | 25 | 200m × 200m | 1 | 1.30 | 5 | N/A | 3 | $8.33 \times 10^4$ | 16.0 (~ 4.41 hrs↓) | 3.9 |
| | 26 | 200m × 200m | 4 | 3.35 | 0 | N/A | 3 | $2.59 \times 10^4$ | 51.8 (~ 7.69 hrs↓) | 2.9 |
| | 27 | 200m × 200m | 4 | 1.76 | 5 | N/A | 3 | $1.79 \times 10^4$ | 49.0 (~ 4.78 hrs↓) | 3.4 |
| Arnold with Index Switching and Map Zoning (Arnold-Logistic) | 28 | 50m × 50m | 1 | N/A | 0 | 1.0 | 1 | $4.70 \times 10^3$ | 32.4 (~ 0.63 hrs↓) | 2.1 |
| | 29 | 50m × 50m | 1 | N/A | 1 | 0.85 | 3 | $5.20 \times 10^3$ | 21.2 (~ 0.39 hrs↓) | 2.8 |
| | 30 | 50m × 50m | 1 | N/A | 4 | 0.15 | 3 | $6.03 \times 10^3$ | 0.3 (~ 0.01 hrs↓) | 3.5 |
| | 31 | 50m × 50m | 1 | N/A | 5 | 0.80 | 1 | $4.35 \times 10^3$ | 58.2 (~ 1.68 hrs↓) | 3.3 |
| | 32 | 200m × 200m | 1 | N/A | 0 | 0.10 | 3 | $8.93 \times 10^4$ | 48.1 (~ 22.97 hrs↓) | 2.5 |
| | 33 | 200m × 200m | 1 | N/A | 5 | 0.25 | 3 | $7.73 \times 10^4$ | 22.1 (~ 6.09 hrs↓) | 3.6 |
| | 34 | 200m × 200m | 4 | N/A | 0 | 0.10 | 3 | $2.28 \times 10^4$ | 57.4 (~ 8.56 hrs↓) | 2.5 |
| | 35 | 200m × 200m | 4 | N/A | 5 | 0.05 | 3 | $1.72 \times 10^4$ | 51.1 (~ 4.98 hrs↓) | 3.3 |
| Arnold with Index Switching and Scaling | 36 | 200m × 200m | 1 | N/A | 0 | N/A | 1 | $1.20 \times 10^5$ | 30.2 (~ 14.4 hrs↓) | 3.3 |
| | 37 | 200m × 200m | 1 | N/A | 5 | N/A | 3 | $6.85 \times 10^4$ | 30.9 (~ 8.53 hrs↓) | 2.3 |
| | 38 | 200m × 200m | 4 | N/A | 0 | N/A | 3 | $2.24 \times 10^4$ | 58.2 (~ 8.67 hrs↓) | 2.5 |
| | 39 | 200m × 200m | 4 | N/A | 5 | N/A | 3 | $1.80 \times 10^4$ | 48.7 (~ 4.75 hrs↓) | 3.4 |
| Arnold with Map Zoning and Scaling | 40 | 200m × 200m | 1 | 1.25 | 0 | 0.25 | 3 | $8.55 \times 10^4$ | 50.3 (~ 24.0 hrs↓) | 2.4 |
| | 41 | 200m × 200m | 1 | 1.75 | 5 | 0.20 | 3 | $6.57 \times 10^4$ | 33.8 (~ 9.31 hrs↓) | 3.1 |
| | 42 | 200m × 200m | 4 | 3.00 | 0 | 1.00 | 3 | $2.00 \times 10^4$ | 62.7 (~ 9.33 hrs↓) | 2.2 |
| | 43 | 200m × 200m | 4 | 1.75 | 5 | 0.50 | 3 | $1.49 \times 10^4$ | 57.5 (~ 5.61 hrs↓) | 2.8 |
| Arnold with Index Switching, Map Zoning, and Scaling | 44 | 200m × 200m | 1 | 1.25 | 0 | 0.25 | 3 | $8.55 \times 10^4$ | 50.3 (~ 24.0 hrs↓) | 2.4 |
| | 45 | 200m × 200m | 1 | 1.75 | 5 | 0.20 | 3 | $6.57 \times 10^4$ | 33.8 (~ 9.31 hrs↓) | 3.1 |
| | 46 | 200m × 200m | 4 | 3.00 | 0 | 1.00 | 3 | $2.00 \times 10^4$ | 62.7 (~ 9.33 hrs↓) | 2.2 |
| | 47 | 200m × 200m | 4 | 1.75 | 5 | 0.50 | 3 | $1.49 \times 10^4$ | 57.5 (~ 5.61 hrs↓) | 2.8 |

In order to verify that the performance ratio is a reasonable measurement for evaluating the proposed techniques, a sweep through Arnold system parameters shows the coverage time on a small obstacle free environment displayed in Fig. 6(a)-(b). The Arnold parameters were in the range: 0.35 to 0.65 for $A$; 0.1 to 0.8 for $B$; and 0.2 to 0.4 for $C$. These ranges are known from bifurcation diagrams to contain regions of desired chaotic behavior for the Arnold system with the initial condition $(x_0, y_0, z_0) = (0, 1, 0)$. The results show that Arnold parameters used in this study for Table 1, as well as [19], are a well performing tuple roughly within the top 10% of parameter combinations tested. As observable in Fig. 6(a), this near-optimality should be taken with caution as the parameter combinations are highly nonlinear in performance as well as highly dependent on different environmental or dynamical system initial conditions. However, the near-optimality shows that the performance ratio provides a good estimate of the performance increase achieved by the novel techniques of this work.

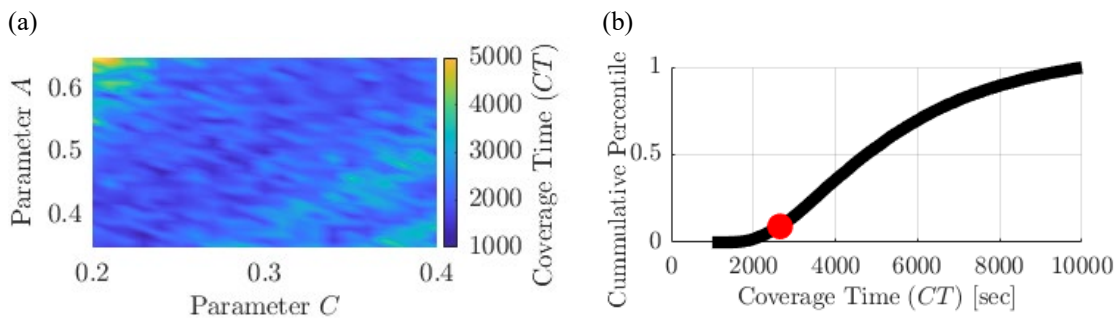

**Fig. 6.** Coverage time of the Original Arnold System for a 50m×50m obstacle free environment (0% occupancy) with SR=4 m. (a) the minimum coverage time over parameter $B$ searched. (b) The cumulative percentile of all Arnold system's parameter combinations achieving 90% coverage in under 10,000 seconds. Red dot indicates $(A, B, C) = (0.5, 0.25, 0.25)$ used for results in Table 1.

### 5.1 Coverage Results in Small Environments

This section discusses the coverage results of the original Arnold system, the Arnold system with the index switching, orientation control technique, and the Arnold-Logistic system in a 50 m × 50 m environment, both with and without obstacles as depicted in Figs.

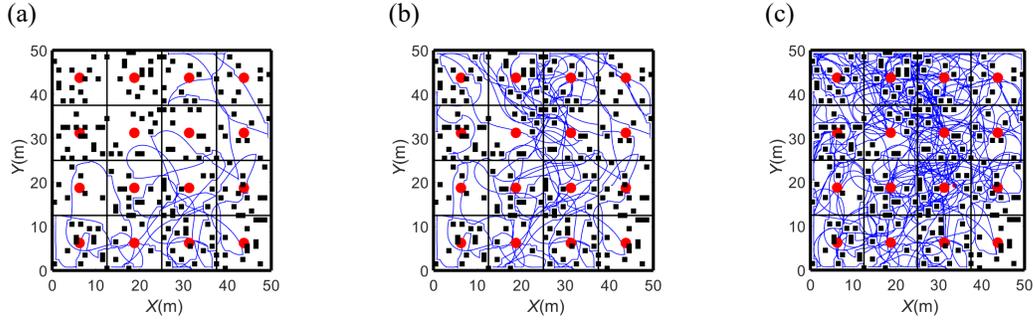

**Fig. 7.** Robot's trajectory with index switching, map zoning, and scaling in a randomly generated, 8% occupancy environment, with: (a) 25% coverage after $1.22 \times 10^3$ seconds, (b) 50% coverage after $3.13 \times 10^3$ seconds, (c) 75% coverage after $9.01 \times 10^3$ seconds. Red dot: zone's midpoint; Black square: obstacle.

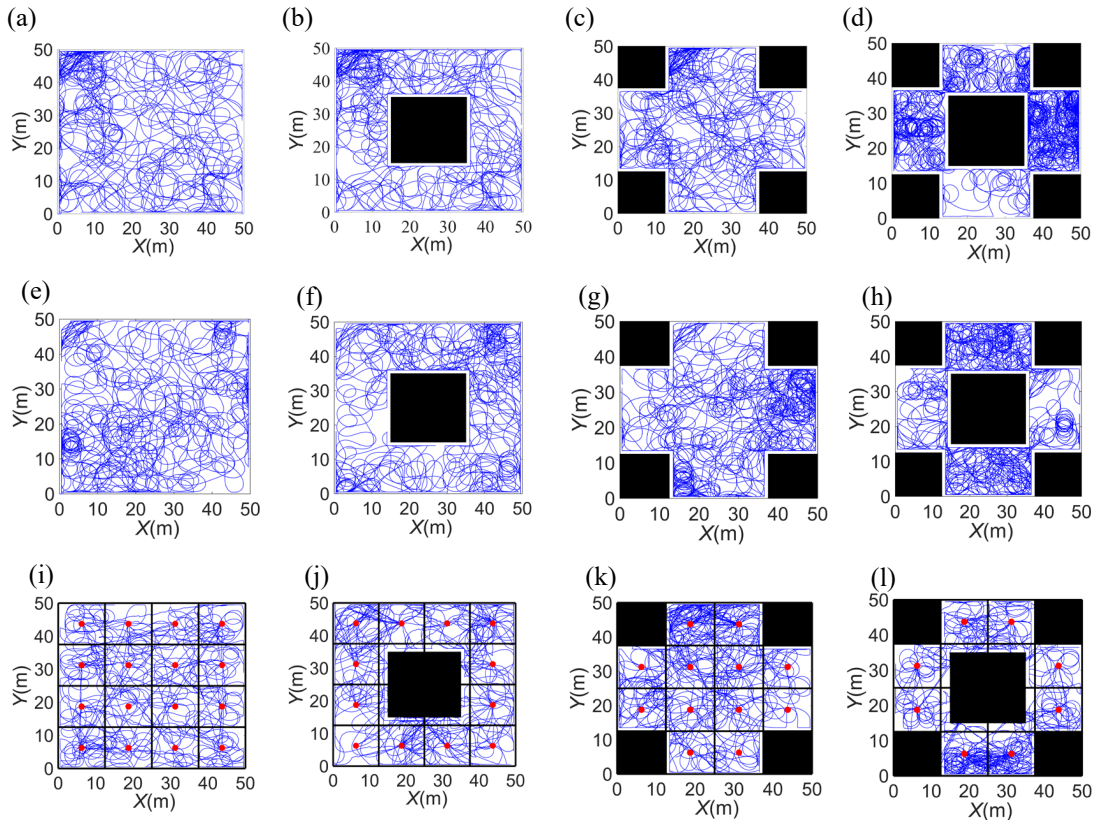

**Fig. 8.** Robot's trajectory for: (a-d) Original Arnold (SR=1 m), (e-h), Arnold with Index Switching (SR=1 m), and (i-l) Arnold Index Switching and Map Zoning (SR=1 m). Dot: zone's midpoint; Black square: obstacle. Obstacle occupancy shown are 0% (a,e,i), 16% (b,f,j), 25% (c,g,k), and 41% (d,h,l).

7 and 8. In all these cases, the robot could successfully cover 90% of the environment while simultaneously avoiding obstacles in its path. Figs. 8(a)-(d) show the robot's trajectories generated using the original Arnold system; dense trajectories can be observed is some parts of the environment, especially in the five obstacles environments. The first chaos control technique, shown in Figs. 8(e)-(h), has mitigated this issue in all of the 50 m × 50 m environments except for the four obstacles environment (Fig.8(g)). The index switching has decreased the coverage time in most scenarios but there are exceptions. In two of the 8 studied cases, it was shown to have degraded the search performance. This is because in certain iterations, the DS keeps generating trajectories that can mostly extend to adjacent regions and changes in $DS_{index}$ did not help the robot to quickly extend the trajectories to more distant areas.

Adding the map zoning technique can help to overcome this issue and guide the robot to navigate to distant regions of the environment. Figs. 8(i)-(l) and Table 1 show the effectiveness of the map zoning technique at decreasing the repetitive coverage and increasing the coverage uniformity in all 50 m × 50 m environment cases, with the exception of the test configuration with four obstacles (case 30) where the improvement was not significant. The greatest improvement corresponds to the environment configuration with five obstacles (case 31) where using the Arnold-Logistic system has led to a 58.2% reduction in coverage time (~1.68 hours) compared to the original Arnold system. As indicated by *PR*, the Arnold-Logistic planner's performance is fairly close to the optimal performance in small obstacle-free spaces.

In addition to the techniques used to reduce the coverage time, this work also introduced various obstacle avoidance algorithms. The effectiveness of these algorithms can be seen not only in regular environments used to generate Table 1 but irregular environments. Figure 7 shows an environment with randomly generated obstacles where the robot employs all three coverage techniques to safely traverse the environment using the Arnold system for sporadic curves as well as the Logistic map for more straight trajectories between zones. There was behavior observed in the random clutter environment of the robot getting temporally trapped within an alcove of the environment with attempting to use the Logistic system to navigate to a different zone. As such, care should be given to choosing the algorithms and techniques proposed in this work for desired deployment environments. The proposed CPP algorithms work best for largely unoccupied environments rather than a complicated maze of narrow corridors. To illustrate, Figs. 7(a)-(c) show some dramatic changes in the robot's heading angle, due to both the obstacle avoidance techniques as well as the map zoning technique. Practically, this means that the system designers should select an appropriate trajectory tracking controller to ensure the sufficient tracking performance for the robot's safety. The demand on the chosen trajectory tracking controller will significantly increase as the occupancy of the environment increases and in the presence of narrow corridors.

### 5.2 Coverage Results in Large Environments

This work shows that system scaling improves performance by adjusting the system trajectories based on the sensing range and the environment size. Table 1 presents the results of the Arnold system performance after adding the system scaling technique in a 200 m × 200 m environment, with zero and five obstacles, traversed by a robot using 1 m and 4 m sensing ranges. Using the system scaling technique has significantly improved the performance in all cases. Comparing the results with those of the Arnold-Logistic system, one observes very close performance in configurations in which the robot has a 4 m sensing range. In the environment with five obstacles and a 1 m sensing range, using the Arnold system with the scaling technique (case 25) led to a significant reduction in the coverage time. The opposite holds for the environment without obstacles and a 1 m sensing range (case 24 versus case 32). Overall, the Arnold system with the scaling technique presented a remarkably beneficial performance, considering that *SystemScaler* corresponds to a comparatively less complex algorithm than *MapZoning*. Therefore, in situations where the computational power is limited, it might be preferable to use the Arnold with the scaling technique instead of a standard Arnold-based chaotic path planner or a chaotic path planner that utilizes either map zoning or index switching techniques.

However, in applications where the computational power is not a concern and a high performance with a low coverage time is the priority, it would be ideal to use the Arnold-Logistic method in conjunction with the scaling technique. As observed in Table 1, the Arnold systems with all three techniques perform better than either the Arnold system with only the scaling technique or the Arnold-Logistic hybrid system in all cases except for case 44 where the percentage of improvement is very close to case 32. Comparing the results of case 45 with those of case 19, 25, and 33 demonstrates the benefit of combining all three techniques, i.e. simultaneous application of all three techniques, can realize a high level of performance that could not be otherwise achieved using any singular technique. Evaluating *PR* in large obstacle-free spaces in Table 1 once again confirms the effectiveness of the any of the techniques proposed in this work for closing the gap between the chaotic planners and the optimal system performance. While each technique was tailored to a different problem encountered, Table 1 shows that the addition of a technique to a chaotic path planner can only improve upon the known performance. Additionally, as it can be seen in Table 1, the performance ratio varies between 2.4 and 3.1 depending on the environment when using all three chaos control techniques. This is satisfactory considering the fact that the main application of the chaotic path planner is for scenarios needing unpredictability in the paths to evade attacks and it is reasonable that some additional time, when compared to the optimal planner, is dedicated to provide the required level of unpredictability.

Figs. 9 and 10 depict the robot trajectories associated with some of the cases presented in Table 1. Fig. 8 compares the robot's trajectories created by the original Arnold system with those generated using the Arnold system with all three techniques in an environment with no obstacles; Figs. 9(a)-(b) and (c)-(d) correspond to the robot with a 1 m sensing range and a 4 m sensing range, respectively. In both cases, using all three techniques significantly reduces the repetitive coverage and improved the search uniformity. Similar conclusions can be drawn from the trajectories presented in Fig. 10 which compares the coverage using the original Arnold system, the Arnold system with the map zoning technique, and the Arnold system with all techniques in an environment with five obstacles. Addition of the scaling technique in Fig. 10(b) has improved the coverage uniformity and reduced the repetitive coverage; but using the map zoning technique in conjunction with the Arnold-Logistic technique can result in a remarkable enhancement as shown in Fig. 10(c). The supplementary video file (Robot trajectory.mp4) shows how the robot advances its trajectories into the different regions of large areas with and without unknown obstacles; these trajectory videos are associated with Figs.9 (b), 9(d), and 10(c).

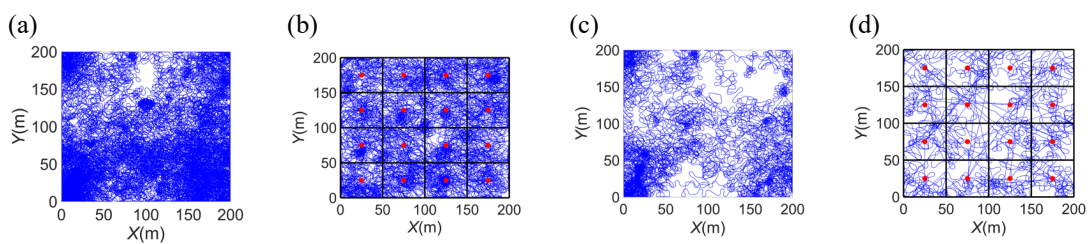

**Fig. 9.** Robot's trajectory in an environment with no obstacles (0% occupancy) for: (a) Original Arnold (SR=1 m), (b) Arnold with Index Switching, Scaling, and Map Zoning (SR=1 m), (c) Original Arnold (SR=4 m), (d) Arnold with Index Switching, Scaling, and Map Zoning (SR=4 m). Dot: zone's midpoint.

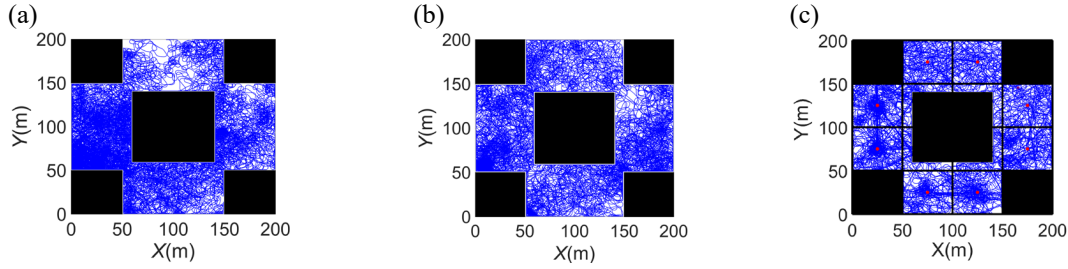

**Fig. 9.** Robot's trajectory in an environment with five obstacles (41% occupancy) for: (a) Original Arnold (SR=1 m), (b) Arnold with Scaling (SR=1 m), (c) Arnold with Index Switching, Scaling, and Map Zoning (SR=1 m). Dot: zone's midpoint; Black square: obstacle.

The decision on whether to use the map zoning technique should be made considering various factors such as the environment estimated size, the estimated density of obstacles in the environment, the properties and coverage density of the generated trajectories by the considered DS, and the robot sensing capabilities. In terms of parameters associated with each technique (e.g., the scaling factor, the initial index, etc.), their effects on the system performance depend on a wide range of factors and cannot be easily predicted. As a result, finding the best parameters is often not a straightforward task in uncertain environments. Next section will discuss how the parameters influence the performance and the methods that can facilitate estimating the optimal parameters in future studies.

### 5.3 Parameters Influence and Future Directions

Table 1 depicts the optimal parameters for each case, obtained through testing a wide range of parameters. Observing the initial index values, the index value of 3 appears to be the optimal choice for almost all the cases except for a few cases (i.e., cases 4, 14, 28, 31, and 36) where index values 1 or 2 were found to be more effective. As far as $c$ is concerned, it depends on the size of the zones. Since we used the same number of zones for all environments, the size of the zones becomes bigger in the larger environments, therefore smaller values of $c$ should suit these environments, as the robot needs to spend more time exploring larger zones. If a large value of $c$ is selected for such cases, the algorithm would force the robot to travel to the other zones too often and potentially result in unnecessary traveling which increases the coverage time. On the other hand, for smaller zones, large values of $c$ should be used (e.g., cases 28-31 in Table 1). However, this expectation does not necessarily hold for all cases in small environments as exceptions may occur (e.g., cases 28 and 30).

Regarding $f$, the direct effect of the environment size, level of obstacle clutter, and sensing range on the selected scaling factor can be clearly observed in cases where only the third control technique is employed (e.g. cases 24-27). In other words, a larger scaling factor is employed in the environments with no obstacle and a robot with high sensing range whereas environments with low clutter and low sensing ranges employ a smaller scaling factor. However, when the other two control techniques become involved (cases 44-47), there is not always an apparent relationship between the scaling factor and the other environmental and platform factors. To illustrate, the optimal scaling factor for case 45 with a SR of 1 m and five obstacles is greater than the scaling factor chosen for case 46 where the SR is 4 m in an obstacle-free environment.

The reasons behind observing inconsistent values for the three parameters discussed above might be the interdependence of different influencing factors and parameters and the complex relationships they have with one another. Future research should perform a more in-depth study of these effects. Moreover, to deal with uncertainties of the environment and the inherent unpredictability of the chaotic path planner, a deep learning-based decision-making strategy appears to be beneficial as it can help determine the required chaos control techniques and optimize their associated parameters in each application given initial knowledge about the environment and the robot's properties.

While this study used a combination of the Arnold and Logistic map systems to realize the second chaos control technique and achieve fast coverage, future studies can examine combinations of other existing continuous systems (e.g., Chen, Rucklidge) and discrete systems (e.g., Taylor-Chirikov map, Hénon map) for the second control technique that would result in paths with different properties and patterns, potentially suited for different applications. Lastly, the algorithms in this study were examined on a two-wheeled differential drive robot; different vehicles correspond to different levels of mobility and maneuverability. The future studies should investigate other types of ground robots or UAVs to evaluate the effects of the robot's characteristics on the path planner's performance.

## 6 CONCLUSIONS

This study proposed new techniques to realize online, fast, and unpredictable search of unknown environments using a mobile robot. Three techniques were established to control the chaotic path planner and enable the robot to automatically cover the environment and adapt to new environments with different sizes and/or present obstacle. The scaling technique provides scalability to the method and allows the robot to adjust the coverage density and extent based on the robot sensing range and environment size. This work also incorporated a new obstacle avoidance technique which allowed the robot to adapt the coverage plan and trajectories when faced with unknown obstacles. The obstacle avoidance technique avoids obstacles efficiently without penalty in the search time performance. Evaluating the search performance of the index switching, map zoning, and scaling techniques in a variety of environments and combinations showed significant improvement in the performance (i.e., reduced coverage time) over a baseline chaotic path planning algorithm for 94% of the tested scenarios. In 2% of cases, negligible improvement in search performance was observed and only the remaining 4% of cases, associated with only utilizing the index switching technique, exhibited decline in the search performance. Combining all three techniques resulted in an average 49% boost in performance within the examined cases. Besides providing unpredictability in the agent's motion and thus enabling it to avoid attacks in adversarial environments, the proposed techniques will ensure a level of performance comparable to that of the optimal path planners. Proper selection of the system's parameters is paramount in ensuring the efficient coverage of the environment. Future studies should develop deep learning-based optimization methods to estimate the parameters of the resulting path planning system based on the known properties of the environment and the robot.